\documentclass[a4paper, 10pt, conference]{IEEEtran}
\IEEEoverridecommandlockouts
\usepackage{cite}
\usepackage{amsmath,amssymb,amsfonts}
\usepackage{algorithm}
\usepackage{algorithmic}
\usepackage{graphicx}
\usepackage{textcomp}
\usepackage{xcolor}
\usepackage{booktabs}
\def\BibTeX{{\rm B\kern-.05em{\sc i\kern-.025em b}\kern-.08em
    T\kern-.1667em\lower.7ex\hbox{E}\kern-.125emX}}
\begin{document}

\title{Sub-optimal Policy Aided Multi-Agent Reinforcement Learning for Flocking Control \\

}

\author{\IEEEauthorblockN{Yunbo Qiu, Yue Jin, Jian Wang, Xudong Zhang}
\IEEEauthorblockA{
Department of Electronic Engineering, Tsinghua University, Beijing, China \\
\{qyb18, jiny16\}@mails.tsinghua.edu.cn, \{jian-wang, zhangxd\}@tsinghua.edu.cn}
}

\IEEEpubid{0000--0000/00\$00.00~\copyright~2022 IEEE}

\maketitle

\begin{abstract}
Flocking control is a challenging problem, where multiple agents, such as drones or vehicles, need to reach a target position while maintaining the flock and avoiding collisions with obstacles and collisions among agents in the environment. Multi-agent reinforcement learning has achieved promising performance in flocking control. However, methods based on traditional reinforcement learning require a considerable number of interactions between agents and the environment. This paper proposes a sub-optimal policy aided multi-agent reinforcement learning algorithm (SPA-MARL) to boost sample efficiency. SPA-MARL directly leverages a prior policy that can be manually designed or solved with a non-learning method to aid agents in learning, where the performance of the policy can be sub-optimal. SPA-MARL recognizes the difference in performance between the sub-optimal policy and itself, and then imitates the sub-optimal policy if the sub-optimal policy is better. We leverage SPA-MARL to solve the flocking control problem. A traditional control method based on artificial potential fields is used to generate a sub-optimal policy. Experiments demonstrate that SPA-MARL can speed up the training process and outperform both the MARL baseline and the used sub-optimal policy.
\end{abstract}

\begin{IEEEkeywords}
flocking control, multi-agent reinforcement learning, sub-optimal policy, artificial potential field
\end{IEEEkeywords}

\section{Introduction}

Flocking control is a challenging multi-agent control problem. In this problem, multiple agents, such as drones or vehicles, need to reach a target position with minimal time and without collisions, while preventing agents from moving far away from others during navigation. Three behaviors that lead to flocking are stated in \cite{reynolds1987flocks}: collision avoidance (avoiding collisions with nearby agents), velocity matching (attempting to match velocity with nearby agents), and flock centering (attempting to stay close to nearby agents).

Many traditional control methods have been proposed to address the problem of flocking control. Leader-follower control methods\cite{yu2016leader,sharma2009flocking} classify agents into leaders and followers. 
Artificial Potential Field (APF) methods\cite{ li2019obstacle,sharma2009flocking} manually construct attractive fields to guide agents to move towards targets and repulsive fields to avoid collisions. Model prediction methods\cite{zhang2015model, lyu2019multivehicle} focus on agents’ ability to predict future movements of other agents.

Methods based on deep reinforcement learning (RL) have been proposed to help address the problem of flocking control more flexibly without a complex and sophisticated design for control signals. Methods based on single-agent RL\cite{wang2018deep, yan2020fixed} consider other agents a part of the environment and learn control policies for each agent independently. However, varying policies of agents during training cause these methods to suffer from a non-stationary environment. Multi-agent RL (MARL) algorithms for flocking control\cite{zhu2020multi, zhao2020research} are thus proposed.

Traditional RL algorithms need a considerable number of samples to improve learning policies’ performance through bootstrapping gradually. The collection of samples can be costly. How to promote sample efficiency becomes a vital issue in RL. One solution is to learn from prior knowledge. 
Imitation learning methods learn from demonstrations generated by experts without online interactions\cite{pomerleau1988alvinn}, or combine learning from demonstrations with online learning\cite{nair2018overcoming, goecks2020integrating}. 
Methods mentioned above are designed mainly for single-agent learning, and they can be extended to multi-agent cases\cite{le2017coordinated, song2018multi}.
However, a large number of expert demonstrations are required to aid agents’ learning in these methods.

As for the problem of flocking control, the APF method has been used to provide prior knowledge to help the training of RL methods. Specifically, reward functions are designed according to APF in\cite{yao2020path}. An action controller combines actions generated by APF and RL to interact with the environment in \cite{li2021path}. Value functions of RL are initialized with potential functions in APF in \cite{cheng2019navigation}. However, these methods did not directly incorporate policies generated by APF into the learning process of RL.

\IEEEpubidadjcol

This paper proposes a sub-optimal policy aided MARL algorithm (SPA-MARL). Instead of learning from the experiences gathered in advance, SPA-MARL directly utilizes a policy that can be manually designed or solved through non-learning methods to aid the process of online MARL training.
Additionally, the performance of the policy can be sub-optimal, which relaxes the rigid requirements for an expert policy performance compared to the imitation learning methods.
Specifically, in SPA-MARL, each agent takes the action generated by the sub-optimal policy as a reference and determines whether to learn from it by comparing it with its own decision. If the action generated by the sub-optimal policy is considered a better one, the agent learns to imitate it while optimizing its online MARL objective. Otherwise, the agent only optimizes its online MARL objective.
Based on our proposed SPA-MARL, we solve the problem of flocking control, where an APF method is used to provide a sub-optimal policy to directly aid MARL, and thus can accelerate learning more efficiently.

The main contributions of this paper are as follows: 
\begin{itemize}
\item A novel algorithm SPA-MARL is proposed to directly learn from a sub-optimal policy by imitating its actions that are evaluated as better ones.
\item The flocking control problem is solved by SPA-MARL with an APF-based sub-optimal policy. Experiments show that SPA-MARL can improve sample efficiency and surpass the performance of both online MARL policy and the sub-optimal policy in flocking control.
\end{itemize}

\section{Background}

\subsection{Markov Games}

In this paper, the problem of flocking control is modeled as Markov games similar to \cite{lowe2017multi}. The Markov games can be represented as a tuple $G=<\mathcal{S},\mathcal{A},\mathcal{T},\mathcal{R},\mathcal{O},n,\gamma>$, where $\mathcal{S}$, $\mathcal{A}$, and $\mathcal{O}$ are the sets of environment states, each agent's actions, and each agent's observations, respectively.

At each time step, each agent simultaneously chooses an action $a_i\in\mathcal{A}$ where $i$ indicates the index of the agent. Actions of all the $n$ agents form a joint action $\boldsymbol{a}\in\mathcal{A}^n$ to interact with the environment. Agents’ actions lead to the transition of the environment state to a new state, represented as $\mathcal{T}: \mathcal{S}\times\mathcal{A}^n\mapsto\mathcal{S}$.
Then, each agent receives a reward $r_i\in\mathcal{R}$. The reward is determined by the environment state and agents' actions $r_i(s,a): \mathcal{S}\times\mathcal{A}^n\mapsto\mathbb{R}$.
Meanwhile, each agent receives a private observation $o_i: \mathcal{S}\mapsto\mathcal{O}$. Each agent's reward and observation form joint reward $\boldsymbol{r}$ and joint observation $\boldsymbol{o}$, respectively.

Each action has a control policy $\pi_i: \mathcal{O}\mapsto\mathcal{A}$ to generate actions. The objective of each agent is to maximize the expectation of its return $R_i=\sum_{t=0}^\infty \gamma^t r_{i,t}$, where $\gamma\in\left(0,1\right]$ is a discount factor. 

\subsection{Multi-Agent Reinforcement Learning Framework}

The actor-critic framework\cite{sutton2018reinforcement} is adopted in many MARL algorithms. ‘Actor’ refers to a policy function $\pi(o_i)$ that generates actions for the agent. ‘Critic’ refers to a value function that estimates the expected return. Action-value function, also called Q-function $Q(o, a)$, is commonly used.

Centralized training and decentralized execution are usually adopted in MARL algorithms\cite{lowe2017multi}. Observations and actions of all the agents are provided for the agent to learn its Q-function during the training process, while extra information about other agents’ observations and actions is no longer provided during the execution process.

MADDPG is a typical actor-critic MARL algorithm for cases of continuous action spaces. In MADDPG, Q-functions and policy functions are parameterized by $\theta_i$ for agent $i$. The loss function of Q-functions is designed as:
\begin{equation}
\begin{aligned}
L_{i, critic}&=\mathbb{E}_{\boldsymbol{o},\boldsymbol{a},r_i,\boldsymbol{o}'\sim \mathcal{D}}[(Q_i (\boldsymbol{o},a_1,...,a_n)-y_{i})^2],\\
\end{aligned}
\label{equ-critic-on}
\end{equation}
where experiences are uniformly sampled from the replay buffer $\mathcal{D}$, and $y_{i}$ is the target for critic optimization:
\begin{equation}
\begin{aligned}
y_{i}&=r_i+\gamma Q'_i (\boldsymbol{o}',a'_1,...,a'_n)|_{a'_j=\pi'_j(o'_j)},\\
\end{aligned}
\label{equ-y-on}
\end{equation}
where $Q'_i$ is the target Q-function of agent $i$, and $\pi'_j$ is the target policy function of agent $j$. Target functions are backups to stabilize the training process.

The loss function of policy functions in MADDPG is designed as:
\begin{equation}
\begin{aligned}
L_{i, actor}&=\mathbb{E}_{\boldsymbol{o},\boldsymbol{a}\sim \mathcal{D}}[-Q_i(\boldsymbol{o},a_1,...,a_i,...,a_n)|_{a_i=\pi_i(o_i)}].\\
\end{aligned}
\label{equ-actor-on}
\end{equation}

\section{Problem Formulation}

\subsection{Spaces of Observations and Actions}

\begin{figure}[t]
\centerline{\includegraphics[width=0.5\textwidth]{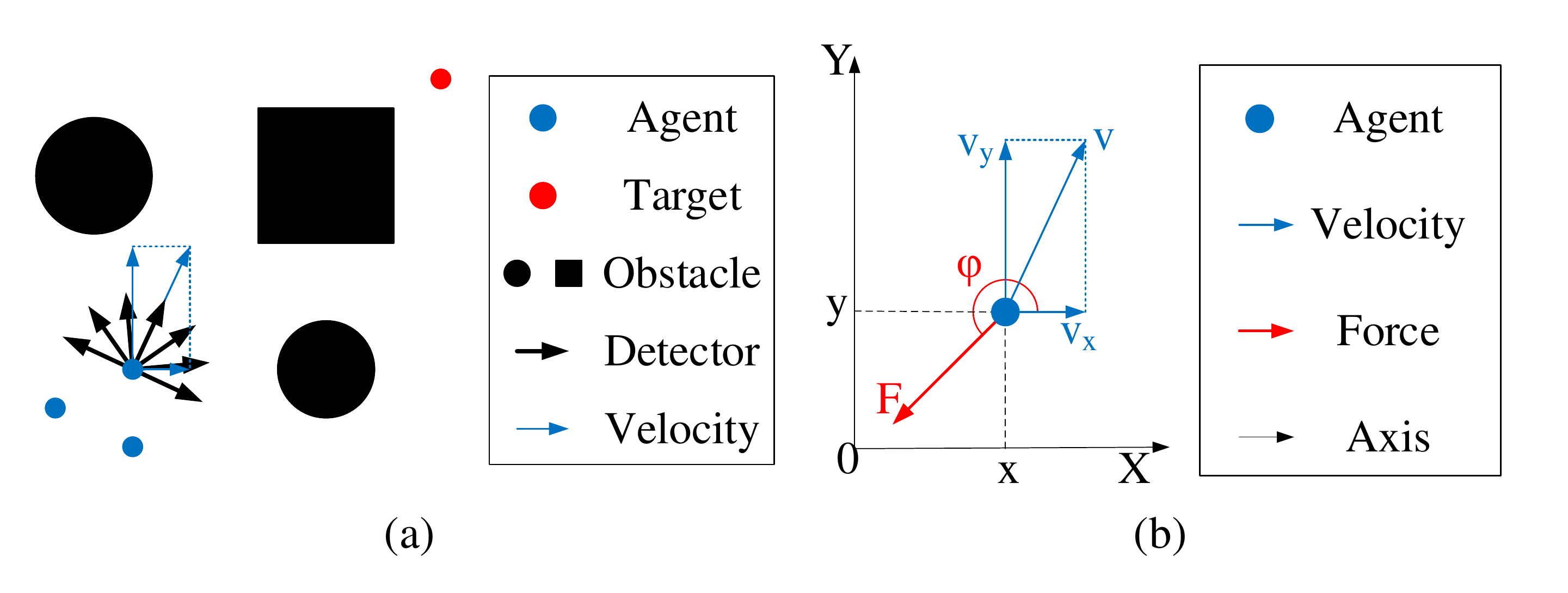}}
\caption{(a)Observations of agents. (b)Actions of agents.}
\label{fig1}
\end{figure}

Observation spaces of agents in MAS are shown in Fig.~\ref{fig1}(a). The observations of each agent consist of:
(1) Relative positions of the target and other agents provided by a global position system.
(2) Obstacle distances detected by 7 rangefinders.
One rangefinder is in the direction where the agent goes forward. Other six rangefinders are evenly distributed on each side of the first rangefinder every 30 degrees. 
The detection range is denoted as the relative distance between the agent and the sensed obstacle in its direction within distance $d_{detect}$.
(3) Horizontal speed $v_x$ and vertical speed $v_y$ of itself.

The actions of each agent are the forces applied to the agents to control agents’ movements, as shown in Fig.~\ref{fig1}(b). The spaces of actions are two-dimensional for magnitude $F$ and angle $\phi$ of the force. The actions control the agents with mass $m$ according to the following dynamic functions:
\begin{equation}
\left\{
\begin{aligned}
&accel=\frac{F}{m},\\
&accel_x=accel\cdot{cos\phi}, accel_y=accel\cdot{sin\phi},\\
&\dot{v_x}=accel_x, \dot{v_y}=accel_y,\\
&\dot{x}=v_x, \dot{y}=v_y,\\
\end{aligned}
\right.
\end{equation}
where $x$ and $y$ are horizontal and vertical coordinates respectively, and $accel$ represents the acceleration.

\subsection{Goal and Reward Functions}

In the flocking control problem, agents are expected to navigate to the target while maintaining the flock. In addition, agents need to avoid collisions with obstacles and collisions among themselves. Fewer time steps and fewer applied forces are also preferred in the task.
An episode of the flocking control task terminates in 3 cases: (1) All the agents arrive near the target, quantified as not being farther from the target than an arrival threshold $d_{arr}$. (2) A collision occurs. (3) The time step of the episode exceeds the maximal episode length $T_{episode}$. An episode succeeds only in case 1.

To guide agents to fulfill each subgoal described above, we design reward functions composed of different terms. Specifically, for an agent $i$ at time step $t$, a transition reward $r_{trans}$ to guide the agent in moving towards the target is defined as: 
\begin{equation}
\begin{aligned}
r_{trans}&=d_{tar} (o_{i,t-1})-d_{tar} (o_{i,t}),\\
\end{aligned}
\end{equation}
where $d_{tar} (o_{i,t})$ denotes the distance between agent $i$ and the target at time $t$, which is a component of observations.

A flocking reward $r_{flock}$ to encourage the agent not to move farther away from the center of the flock than $th_{f}$ is defined as:
\begin{equation}
\begin{aligned}
r_{flock}&=relu(d_f (o_{i,t-1})-th_{f})-relu(d_f (o_{i,t})-th_{f}),\\
\end{aligned}
\end{equation}
where $d_f (o_{i,t})$ denotes the distance between agent $i$ and the center of the flock at time $t$,  and $relu(x)=max(0,x)$ is the linear rectification function. 

A collision reward $r_{col}$ is defined as:
\begin{equation}
r_{col}=
\left\{
\begin{aligned}
&d_{obs} (o_{i,t})-th_{col},&&{\text{if }}d_{obs} (o_{i,t})<th_{col},\\
&0,&&{\text{else }},\\
\end{aligned}
\right.
\end{equation}
where $d_{obs} (o_{i,t})$ denotes the distance sensed by rangefinders between agent $i$ and the nearest obstacle at time $t$. It penalizes the agent if the agent is nearer to an obstacle than a distance threshold $th_{col}$, and thus instructs the agent to avoid collisions with obstacles. 

A crossing reward $r_{cross}$ is defined as:
\begin{equation}
r_{cross}=
\left\{
\begin{aligned}
&d_{ag} (o_{i,t},j)-th_{cross},&&{\text{if }}d_{ag} (o_{i,t},j)<th_{cross},\\
&0,&&{\text{else }},\\
\end{aligned}
\right.
\end{equation}
where $d_{ag} (o_{i,t},j)$ denotes the distance between agent $i$ and another agent $j$ at time $t$. It penalizes the agent if the agent is closer to another agent than a distance threshold $th_{cross}$, and thus instructs the agent to avoid collisions with others.

A step reward $r_{step}$ to urge the agent to complete the task as soon as possible is defined as:
\begin{equation}
\begin{aligned}
r_{step}&=-1.\\
\end{aligned}
\end{equation}

A stability reward $r_{stab}$ to encourage the agent to complete the task with as least change in velocity as possible is:
\begin{equation}
\begin{aligned}
r_{stab}&=-\left| F\right|.\\
\end{aligned}
\end{equation}

The integrated reward $r_{i,t}$ for an agent $i$ in time step $t$ is a linear combination of the rewards mentioned above:
\begin{equation}
\begin{aligned}
r_{i,t}=&\rho_{trans}\cdot{r_{trans}}+\rho_{flock}\cdot{r_{flock}}+\rho_{col}\cdot{r_{col}}\\
&+\rho_{cross}\cdot{r_{cross}}+\rho_{step}\cdot{r_{step}}+\rho_{stab}\cdot{r_{stab}},\\
\end{aligned}
\end{equation}
where $\rho_{trans}$, $\rho_{flock}$, $\rho_{col}$, $\rho_{cross}$, $\rho_{step}$, and $\rho_{stab}$ are coefficients of corresponding rewards.

\section{Algorithm}

\subsection{Sub-optimal Policy Aided MARL}

A sub-optimal policy aided MARL algorithm (SPA-MARL) is proposed in this paper for flocking control. SPA-MARL can base on an actor-critic MARL algorithm. In this paper, we build SPA-MARL upon MADDPG.

SPA-MARL consists of an online learning part similar to MADDPG and a sub-optimal policy aided part. During the training of online MARL, parameters of Q-functions and policy functions are initialized randomly. Therefore, agents demand a lot of samples of experiences to optimize Q-functions and policy functions gradually. It requires agents to continuously interact with the environment, which is of low sample efficiency. Using a sub-optimal policy $\mu$ easily designed with prior knowledge to aid the process of online MARL training is thus the focus of our algorithm. It mainly aims to improve sample efficiency of training. 

Although the sub-optimal policy is not the best policy for the agents, it outperforms the randomly initialized policy by a large margin. Thus, agents are expected to learn from the sub-optimal policy, especially in the early stage of training. Accordingly, the policy $\pi_i$ is optimized by minimizing the following loss:
\begin{equation}
\begin{aligned}
L_{i,actor,ac}&=\mathbb{E}_{\boldsymbol{o},\boldsymbol{a}\sim \mathcal{D}}[(\pi_i(o_i)-\mu(o_i))^2],\\
\end{aligned}
\label{actionloss}
\end{equation}
where $\mu(o_i)$ is the sub-optimal policy with observation $o_i$ as its input. By minimizing this loss, agents learn to generate actions as similar to the sub-optimal policy as possible. 

Note that the loss function in (\ref{actionloss}) involves the action differences over the samples collected from agents' online interactions with the environment. Therefore, it is different from behavior cloning (BC) based methods \cite{pomerleau1988alvinn, nair2018overcoming,goecks2020integrating} and learning from offline data \cite{fujimoto2021minimalist}. These methods use data collected by experts or other existing policies in advance to compute action differences. In addition to requiring an extra dataset, these methods only imitate limited samples from the fixed dataset. In contrast, SPA-MARL imitates the actions given by the sub-optimal policy at every state encountered by agents, and thus can guide agents' learning more directly and accurately. To distinguish our method from BC-based methods, we name the loss in (\ref{actionloss}) action cloning loss.

Since the sub-optimal policy is not optimal, the online MARL part is still expected to improve the performance of learning policies. The whole loss for policy functions combines the online MARL loss as (\ref{equ-actor-on}) and the action cloning loss that encourages the learning policy to act like the sub-optimal policy: 
\begin{equation}
\begin{aligned}
L_{i,actor}^{naiveSPA}&=L_{i,actor}+\alpha_{i}\cdot L_{i,ac},\\
\end{aligned}
\label{equ-actor}
\end{equation}
where $\alpha$ is a parameter deciding the mix ratio of online MARL loss and action cloning loss.

To automatically balance the mix ratio of the two losses, $\alpha_i$ is calculated similar to \cite{fujimoto2021minimalist} as:
\begin{equation}
\begin{aligned}
\alpha_{i}&=\frac{1}{M}\sum_{\boldsymbol{o},\boldsymbol{a}\sim \mathcal{M}}|Q_i(\boldsymbol{o},a_1,...,a_i,...,a_n)|_{a_i=\pi_i(o_i)}|.\\
\end{aligned}
\end{equation}
Its purpose is to transform the action cloning loss and online MARL loss to the same scale. $\alpha_{i}$ is calculated for an agent $i$ in every mini-batch with $M$ experiences.

In the training process, as the learning policy’s performance gradually approaches the sub-optimal policy, the sub-optimal policy provides less help for training, and may even hinder the learning policy from achieving better performance. Therefore, an additional term is added to the action cloning loss to form dynamic action cloning loss for policy functions:
\begin{equation}
\begin{aligned}
L_{i,dac}=&\mathbb{E}_{\boldsymbol{o},\boldsymbol{a}\sim \mathcal{D}}[(\pi_i(o_i)-\mu(o_i))^2\\
&\cdot \epsilon(Q_i(\boldsymbol{o},a_1,...,a_n)|_{a_j=\mu(o_j)} \\
&- Q_i(\boldsymbol{o},a_1,...,a_i,...,a_n)|_{a_i=\pi_i(o_i)})],\\
\end{aligned}
\label{equ-actor-spa}
\end{equation}
where $\epsilon(x)=max(0,\frac{x}{|x|})$ is a step function. It only chooses the experiences where the values of Q-functions of the actions generated by the sub-optimal policy are higher than those corresponding to the actions generated by the learning policy. In this case, the sub-optimal policy can obtain higher expected return than the learning policy if Q-function is accurate enough. Actually, bootstrapping in RL continues optimizing Q-function to be more accurate. Therefore, as learning proceeds, the loss in (\ref{equ-actor-spa}) can recognize the really better actions. With this adaption, the sub-optimal policy will not harm the learning policy's performance by enforcing the learning policy to act like the sub-optimal policy at every state.

Accordingly, the whole loss for policy functions is adapted to take advantage of the dynamic action cloning loss as:
\begin{equation}
\begin{aligned}
L_{i,actor}^{SPA}&=L_{i,actor}+\alpha_{i}\cdot L_{i,dac},\\
\end{aligned}
\label{equ-actor-spawhole}
\end{equation}

As for the critics of agents, the target $y_i$ for critic optimization in (\ref{equ-critic-on}) is also modified to optimize Q-functions with the help of the sub-optimal policy:  
\begin{equation}
\begin{aligned}
y_{i}^{SPA}=&r_i+\gamma  max(Q'_i (\boldsymbol{o}',a'_1,...,a'_n)|_{a'_j=\pi'_j(o'_j)},\\
& Q'_i (\boldsymbol{o}',a'_1,...,a'_i,...,a'_n)|_{a'_i=\mu(o'_i),a'_{j\neq{i}}=\pi'_j(o'_j)}),\\
\end{aligned}
\label{equ-y-spa}
\end{equation}
which compares target Q-function of target policy with that of the sub-optimal policy. It aims to compare the next action’s expected returns of the learning policy and the sub-optimal policy. If actions generated by the sub-optimal policy obtain a higher value of action-value function, it instructs the learning Q-function to know a potentially better transition for future. Q-functions can accordingly select a better transition for future to be updated. It thus aids Q-functions in estimating expected returns better and accelerating the training process.

The whole algorithm is summarized in Algorithm \ref{SPA-MARL}.

\begin{algorithm} [t]
	\caption{SPA-MARL} 
	\label{SPA-MARL} 
	\begin{algorithmic}
	    \REQUIRE a sub-optimal policy $\mu$
	    \FOR {episode $= 1 \text{ to } E$ }
	    \STATE Initialize a random process $\mathcal{N}$ for action exploration
        \STATE Receive initial observation $\boldsymbol{o}$
        \FOR{$t$ = 1 to $T_{episode}$}
        \STATE for each agent $i$, select action $a_i = \pi_{\theta_i}(o_i) + \mathcal{N}_t$ w.r.t. the current policy and exploration
        \STATE Execute actions $\boldsymbol{a} = (a_1, . . . , a_n )$ and observe reward $r$ and new observation $\boldsymbol{o}'$
        \STATE Store $(\boldsymbol{o}, \boldsymbol{a}, \boldsymbol{r}, \boldsymbol{o}')$ in replay buffer $\mathcal{D}$
        \STATE $\boldsymbol{o} \leftarrow \boldsymbol{o}'$
	    \FOR {agent $i = 1 \text{ to } n$ }
	    \STATE Randomly sample a mini-batch $M$ from $\mathcal{D}$
	    \STATE Update actor by minimizing (\ref{equ-actor-spawhole})
        \STATE Update critic by minimizing (\ref{equ-critic-on}) with (\ref{equ-y-spa}) as target
	    \ENDFOR
	    \STATE Update target networks: $\theta '_i \leftarrow \tau \theta _i + (1-\tau) \theta '_i$
	    \ENDFOR
	    \ENDFOR
	\end{algorithmic} 
\end{algorithm}

\subsection{Artificial Potential Field for Sub-optimal Policy}

To apply SPA-MARL to solve the flocking control problem, we adopt the artificial potential field (APF) method to generate a policy as a prior sub-optimal policy. 
The APF method is a traditional method for flocking control. It constructs potential fields for every agent according to subgoals of the task. Forces applied to control the movement of an agent are the derivatives of these potential fields correspondingly.

Traditional APF methods\cite{li2019obstacle,sharma2009flocking} have global information about locations of obstacles. This paper considers that only local detection of rangefinders is known.
Therefore, potential fields of the sub-optimal policy are updated in real time based on current observations.

Specifically, for an agent $i$ at time step $t$, the attractive potential field for navigation to the target is designed as:
\begin{equation}
U_{nav} = 
\left\{
\begin{aligned}
&\frac{1}{2} d_{tar}(o_{i,t})^2,&&{\text{if }}d_{tar}(o_{i,t})<d_{arr},\\
&d_{arr} \cdot d_{tar}(o_{i,t}) - \frac{1}{2} d_{arr}^2 ,&&{\text{else }}.\\
\end{aligned}
\right.
\end{equation}

The attractive potential field for flocking is designed as:
\begin{equation}
U_{flock} = 
\left\{
\begin{aligned}
&d_{f}\cdot d_{f}(o_{i,t}),&&{\text{if }}d_{f}(o_{i,t})>d_{f},\\
&d_{f},&&{\text{else }},\\
\end{aligned}
\right.
\end{equation}
where $d_{f}$ is the safe distance between an agent and the center of the flock. 

The repulsive potential field for avoiding collisions between the agent and obstacles is called collision potential field. Each rangefinder $k$ constructs a particular collision potential function as:
\begin{equation}
U_{col,k} = 
\left\{
\begin{aligned}
&\frac{1-\delta}{2}(\frac{1}{d_{obs}(o_{i,t},k)}-\frac{1}{d_{col}})^2, \\
& \quad \quad \quad \quad{\text{if }}d_{obs}(o_{i,t},k)<d_{col},\\
&0 ,\quad \quad \quad{\text{else }},\\
\end{aligned}
\right .
\end{equation}
where $d_{obs}(o_{i,t},k)$ is the distance between the agent $i$ and the nearest obstacle detected by rangefinder $k$, and $d_{col}$ is the safe distance to avoid collisions.
$\delta$ is defined as:
$\delta = 0.2\times\frac{\Delta_{det}}{30}$, 
where $\Delta_{det}$ means the difference in degrees of the rangefinder that senses the obstacle and the rangefinder in the direction of moving. 

The repulsive potential field for avoiding collisions among agents is called crossing potential field. Each other agent $j$ results in a crossing potential function as:
\begin{equation}
U_{cross,j} = 
\left\{
\begin{aligned}
&(\frac{1}{d_{ag}(o_{i,t},j)}-\frac{1}{d_{cross}})^2, \\
&\quad \quad \quad \quad{\text{if }}d_{ag}(o_{i,t},j)<d_{cross},\\
&0 ,\quad \quad \quad{\text{else }},\\
\end{aligned}
\right .
\end{equation}
where $d_{cross}$ is the safe distance to avoid collisions among agents.

$F_{nav}$, $F_{flock}$, $F_{col,k}$, and $F_{cross,j}$ are the derivatives of these potential fields correspondingly. The joint force applied to an agent is the linear combination of four types of forces:
\begin{equation}
\begin{aligned}
F_{i,t} =&\beta_{nav}\cdot F_{nav}+\beta_{flock}\cdot  F_{flock}\\
&+\beta_{col}\cdot \sum_{k} F_{col,k}+ \beta_{cross} \cdot \sum_{j} F_{cross,j},\\
\end{aligned}
\end{equation}
where collision forces are summed over 7 rangefinders, and crossing forces are summed over other $n-1$ agents. Agents' actions of the APF method, including magnitude and angle of applied forces, are generated according to joint forces.

\section{Experiments}

\subsection{Experiment Settings}

An environment for flocking control is built to verify SPA-MARL, as in Fig.~\ref{fig4}. The environment map is square, whose side length is $G$ $unit$. Half of $m$ obstacles are round, while the others are square. At the start of each episode, $n$ agents are close to each other. Agents are round in shape with radius $D_{agent}$, and their mass is 1. The target is square with side length $D_{target}$, located near the edge of the map. The size of obstacles and the positions of the target, agents, and obstacles are randomly initialized in every episode.

In our experiments, $G$ is set as 36 and $unit$ is 20 for the map. $n$ is 3, and $m$ is 10. $D_{agent}$ is 0.2 $unit$ and $D_{target}$ is 0.15 $unit$. The diameter or side length of obstacles are uniformly randomized from 3 to 5 $unit$. For rangefinders, $d_{detect}$ is 4 $unit$. Arrival threshold $d_{arr}$ is $n ~ unit$. 
Maximal acceleration of agents is 0.5 $unit$, and maximal speed of agents is 0.5 $unit$.

For Markov games, $\gamma$ is 0.95 and maximal episode length $T_{episode}$ is 100 time steps. For the reward scheme, $\rho_{trans}$ is $\frac{0.25}{G}$, $\rho_{flock}$ is $\frac{0.5}{G}$, $\rho_{col}$ is $\frac{40}{G}$, $\rho_{cross}$ is $\frac{20}{G}$, $\rho_{step}$ is $\frac{1}{G}$, and $\rho_{stab}$ is $\frac{1}{G}$. Thresholds are $\frac{n}{2}$ $unit$ for $th_{f}$, 0.5 $unit$ for $th_{col}$, 1 $unit$ for $th_{cross}$.

For the APF method, $d_{f}$, $d_{col}$, and $d_{cross}$ are all 1 $unit$. $\beta_{nav}$ is $\frac{1}{d_{arr}}$, $\beta_{flock}$ is $\frac{0.1}{d_{f}}$, $\beta_{col}$ is $\frac{5}{unit^3}$, and $\beta_{cross}$ is $\frac{5}{unit^3}$. 

Neural networks representing policy functions and Q-functions are fully connected networks with $tanh$ as activation function. Each neural network has 3 hidden layers with 64 units per layer. We use Adam to optimize the parameters, with a learning rate of 0.001. Target networks are softly updated with $\tau= 0.0004$. The capacity of replay buffer $\mathcal{D}$ is 300000, and the mini-batch size $M$ is 32. 
Parameters are updated every 4 time steps in training. Agents are trained over $E=200000$ episodes. Results of different algorithms are all run in 3 seeds.

\subsection{Main Results}

\begin{figure}[t]
\centerline{\includegraphics[width=0.5\textwidth]{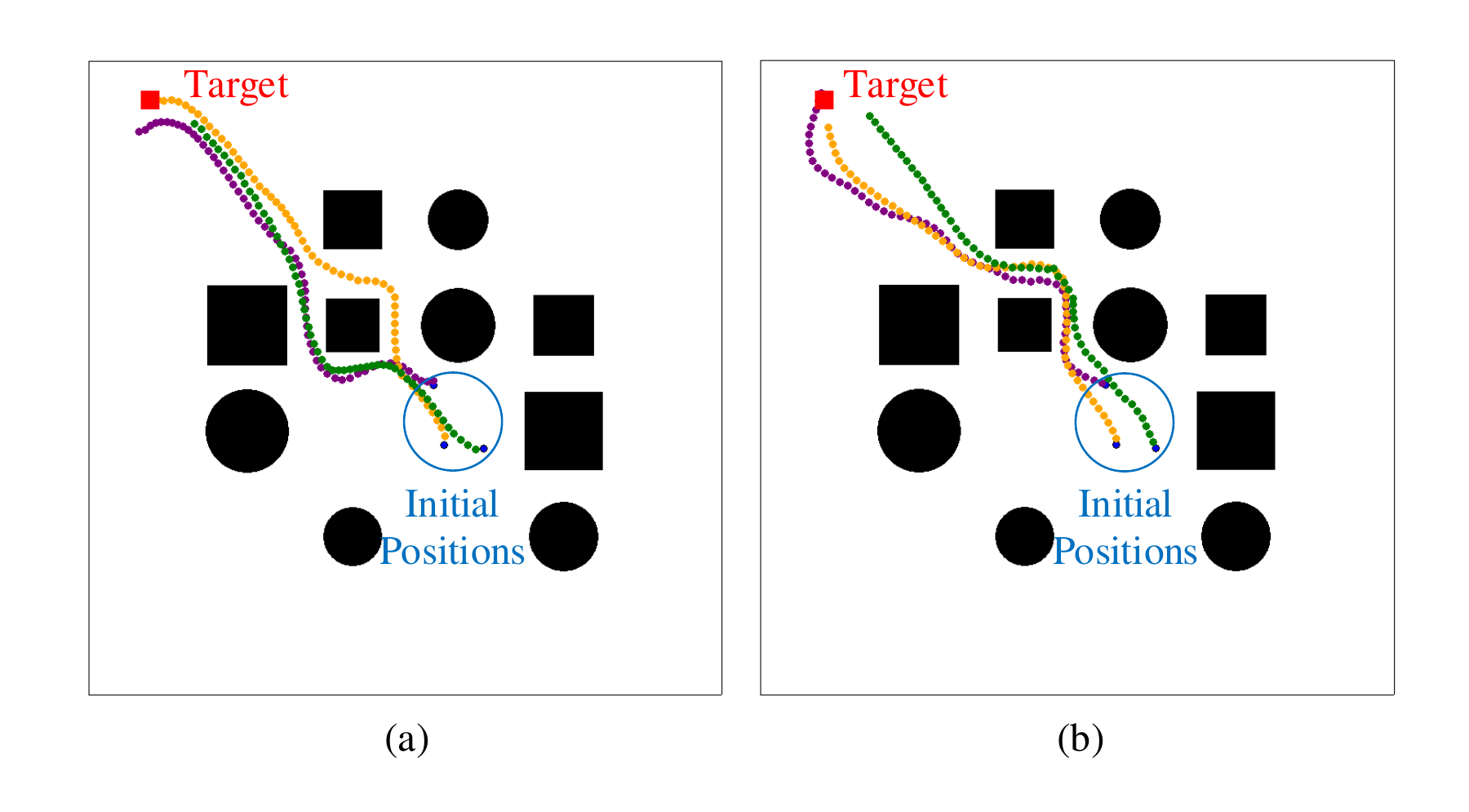}}
\caption{Environment and trajectories of an episode. Obstacles are black. The target is red, and it is enlarged in this figure. The initial positions of agents are blue, and they are circled in this figure. Trajectories of the three agents are green, purple, and orange, respectively. (a)Trajectories of MADDPG. (b) Trajectories of SPA-MARL.}
\label{fig4}
\end{figure}

\begin{table*}[t]
\caption{Statistical Results}
\begin{center}
\begin{tabular}{lccccccc}
\toprule
\textbf{Algorithm}                         & \textbf{Success Rate} & \textbf{Reward} & \textbf{Collision Rate} & \textbf{Crossing Rate} & \textbf{Flock Distance} & \textbf{Time Steps} &\textbf{Force} \\ \midrule
\textbf{SPA-MARL(ours)}        & 0.890                 & 1.031          & 0.055                   & 0.053               & 1.640                   & 40.115                             & 45.792                             \\
\textbf{MADDPG}          & 0.857                 & 0.823          & 0.088                   & 0.045               & 1.702                   & 44.085                             & 33.850                             \\
\textbf{BC}              & 0.392                 & -2.917         & 0.074                   & 0.532               & 1.578                   & 26.019                             & 37.643                             \\
\textbf{Sub-optimal}      & 0.803                 & -0.364         & 0.028                   & 0.140               & 1.807                   & 44.546                             & 62.912                             \\
\textbf{SPA-MARL-actor(ablation)}  & 0.885                 & 0.985          & 0.059                   & 0.054               & 1.635                   & 39.774                             & 45.480                             \\
\textbf{SPA-MARL-critic(ablation)} & 0.875                 & 0.950          & 0.073                   & 0.048               & 1.665                   & 44.544                             & 32.305                             \\ \bottomrule
\end{tabular}
\label{tab1}
\end{center}
\end{table*}

\begin{figure}[t]
\centerline{\includegraphics[width=0.5\textwidth]{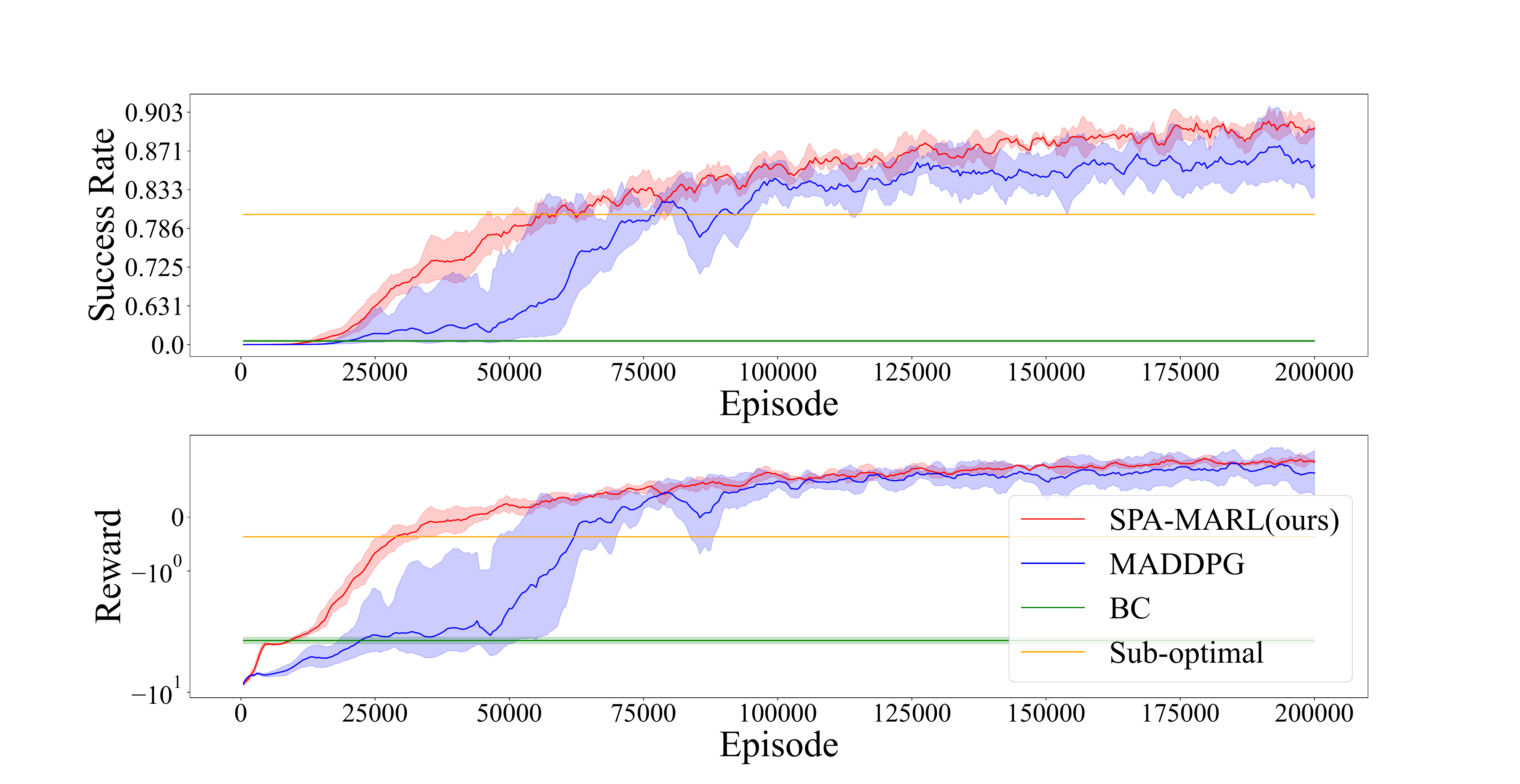}}
\caption{Convergence curves of success rate and reward of main experiments, presented in the fifth root scale and the symmetric log scale, respectively.}
\label{fig6}
\end{figure}

To verify our proposed algorithm SPA-MARL in flocking control, we contrast SPA-MARL with  MADDPG\cite{lowe2017multi} and behavior cloning (BC)\cite{pomerleau1988alvinn}. In addition, the performance of the sub-optimal policy generated by the APF method is also presented. MADDPG is a MARL algorithm, adopted as the MARL part of SPA-MARL. BC learns from demonstrations generated by the sub-optimal policy. In BC, experiences of 8000 episodes are stored in a buffer with the same capacity as SPA-MARL. Agents in BC have been trained 1000000 iterations. Other settings and parameters in MADDPG and BC are the same as SPA-MARL.

The convergence curves of success rate and reward are plotted in Fig.~\ref{fig6}.
As can be seen from the curves, SPA-MARL learns faster and achieves a higher success rate and a higher reward than MADDPG, which demonstrates the superiority of SPA-MARL in sample efficiency.
In addition, SPA-MARL outperforms the sub-optimal policy and BC by a large margin. It indicates SPA-MARL can surpass both the policy that it learns from and the traditional BC method.

Apart from the convergence performance, we also test the performance of the learned flocking control policies of different algorithms with 2500 randomly generated environments. Statistical results are presented in Table \ref{tab1}. 
We evaluate the performance of flocking control with seven metrics. Specifically, ‘Flock Distance’ means the average distance between agents and the center of the flock. ‘Time Steps’ means the average time steps for an episode. ‘Force’ means the average magnitude of forces applied to agents for an episode. 

As in Table \ref{tab1}, SPA-MARL surpasses MADDPG, BC, and the sub-optimal policy in success rate and reward. The performance of BC is inferior largely to the sub-optimal policy, since it merely learns from the sub-optimal demonstrations. As for flock distance and time steps, SPA-MARL is also better than MADDPG and the sub-optimal policy, at the cost of more forces to control agents more precisely.

Furthermore, the sub-optimal policy has a high crossing rate and a low collision rate, which helps SPA-MARL elevate its success rate mainly by decreasing the collision rate compared to MADDPG. Correspondingly, the crossing rate of SPA-MARL is slightly higher than MADDPG under the influence of the sub-optimal policy.

An example of trajectories of SPA-MARL and MADDPG is plotted in Fig.~\ref{fig4}. 
The trajectories demonstrate that SPA-MARL can maintain the flock more tightly during navigation.

\subsection{Ablation Results}

\begin{figure}[t]
\centerline{\includegraphics[width=0.5\textwidth]{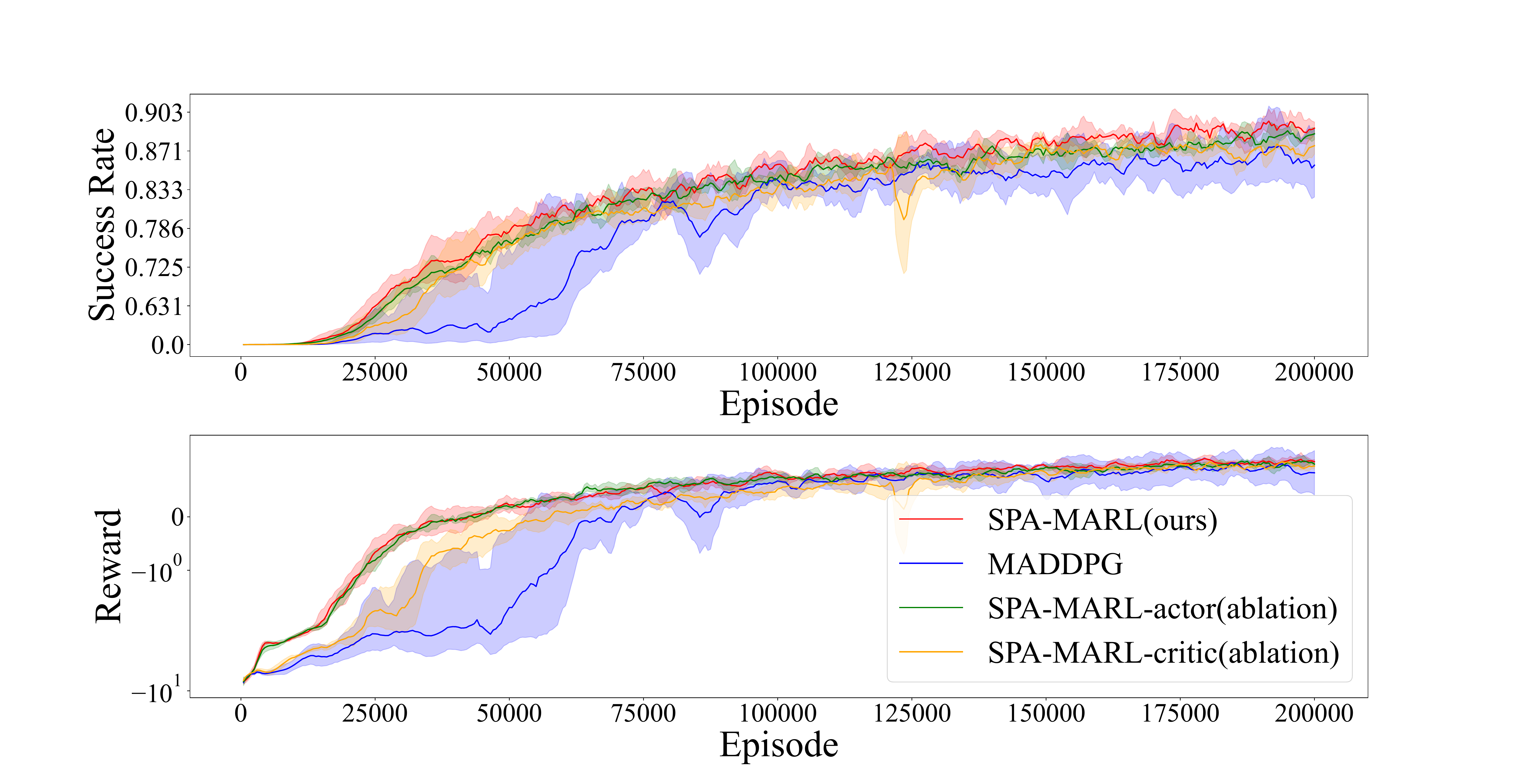}}
\caption{Convergence curves of success rate and reward of ablation experiments, presented in the fifth root scale and the symmetric log scale, respectively.}
\label{fig7}
\end{figure}

To evaluate the effect of the actor part and the critic part of SPA-MARL, two algorithms are designed with only the actor part's adaption and only the critic part's adaption based on MADDPG, which are called SPA-MARL-actor and SPA-MARL-critic, respectively.

The convergence curve of success rate is plotted in Fig.~\ref{fig7}. 
It can be seen that both SPA-MARL-actor and SPA-MARL-critic outperform MADDPG in sample efficiency. In addition, SPA-MARL has better performance in sample efficiency than SPA-MARL-actor and SPA-MARL-critic. Although the convergence speed of SPA-MARL-critic almost catches up to that of SPA-MARL, SPA-MARL still achieves a higher success rate and reward than the two algorithms.

Statistical results of the ablation algorithms are shown in the bottom two lines in Table \ref{tab1}.
The performance in success rate and reward of the two ablation algorithms surpasses MADDPG, while it is inferior to the performance of SPA-MARL. 
Compared with MADDPG, SPA-MARL-actor can complete the task in fewer time steps but requires more forces to control. Such influence of adaption of actor part can be seen in the comparison of SPA-MARL-critic and SPA-MARL as well. Therefore, ablation experiments verify that both adaptions of the actor part and the critic part in SPA-MARL assist in improving success rate, and the actor part aids agents in completing tasks more quickly.

\section{Conclusion and Future Work}

An algorithm named SPA-MARL is proposed for flocking control in this paper. 
SPA-MARL directly leverages a sub-optimal policy to get a reference and decides whether to learn from the sub-optimal policy automatically. The sub-optimal policy aids SPA-MARL in improving sample efficiency and final performance compared to online MARL algorithms.
A sub-optimal policy is solved by the artificial potential field method in flocking control. Experimental results demonstrate that SPA-MARL outperforms both the baseline MARL algorithm and the used sub-optimal policy in learning speed, success rate, flock distance, and navigation time.

In the near future, we expect to study the scalability of SPA-MARL with more agents. Besides, SPA-MARL that employs other sub-optimal policies or other MARL algorithms deserves attempting, which is also left for future work.

\bibliographystyle{IEEEtran}
\bibliography{ref}

\end{document}